\documentclass{article}

\PassOptionsToPackage{numbers, compress}{natbib}
\usepackage[preprint]{nips_2018}

\usepackage{url}
\usepackage{authblk}
\usepackage{alltt}
\usepackage{xcolor}
\usepackage{fancyhdr}
\usepackage{hyperref}
\usepackage{url}
\usepackage{bm}
\usepackage{graphicx}
\usepackage{subcaption}
\usepackage{algorithm}
\usepackage{algorithmic}
\usepackage{amsmath}
\usepackage{listings}
\title{Zap: Making Predictions Based on Online User Behavior \\[.1em]
}

\author{
\small Yuri Chervonyi, Dragos Harabor, Brian Zhang, Josh Sacks \\
\texttt{\{yuri\thanks{To whom correspondence should be addressed: yuri@bytegain.com.}, dragos, brian, jjs\}@bytegain.com}
}

\date{\today}

\begin{document}

\maketitle

\begin{abstract}

\begin{small}

This paper introduces Zap, a generic machine learning pipeline for making predictions based on online user behavior. 
Zap combines well known techniques for processing sequential data with more obscure techniques such as Bloom filters, bucketing, and model calibration into an end-to-end solution. 
The pipeline creates website- and task-specific models \textit{without knowing anything about the structure of the website}. 
It is designed to minimize the amount of website-specific code, which is realized by factoring all website-specific logic into example generators. 
New example generators can typically be written up in a few lines of code.

\vspace{5mm}

\end{small}

\end{abstract}

\section{Introduction}

As more people spend their days on the Internet, businesses have followed by exchanging their physical stores for websites and apps. 
However, in the process of moving online, many businesses have lost the personal touch they had with their customers. 
Instead, they serve generic experiences and random pop-ups. 
The irony is that the increase in online activity means that user web data\footnote{By web data we mean user action on websites or mobile applications.} has evolved into a high-resolution reflection of a person's daily life. 
Businesses could use this highly textured online data to predict user behavior, provide recommendations and increase engagement. 
A recent study commissioned by the Digital Advertising Alliance (DAA) finds that over 40\% of respondents would prefer personalized marketing based on their online behavior \cite{ref:behavioral-marketing-study}. 
Many consumers have also experienced, and now expect, the magic of Google Search autocomplete, Netflix recommendations and more. 
Every business, not just big tech companies, could benefit from personalized experiences. 
Moving forward, every business, not just big tech companies, could benefit from personalized experiences. 

Analyzing web data is a complicated problem and machine learning has become a standard method of solving such problems when heuristics become too complex. 
Machine learning becomes especially effective when a large amount of data is available in which case deep learning (for introduction see \cite{ref:deep-learning-book}) is almost a canonical approach producing state-of-the-art results. 

Most modern deep learning models are based on artificial neural networks\footnote{We will drop ``artificial" from now on.}. 
A neural network is a collection of simple elements called neurons. 
Neurons receive input, change their internal state (activation) and produce output based on the input and activation. 
The network is formed by connecting the output of certain neurons to the inputs of other neurons. 
The strength of connections between neurons is encoded via weights that can be modified by a process called learning. 

We believe that neural networks represent the beginning of a fundamental shift in software development \cite{ref:software20}. 
In ``classic" software development (Software 1.0), the programmer identifies a specific point in program space with some desirable behavior and expresses that by writing lines of code. 
In contrast, neural networks (Software 2.0) are written in a more abstract language, such as the weights of a neural network. 
The weights are not explicitly specified but learned via backpropagation and stochastic gradient descent. 

Deep learning has made advances in solving problems that have previously resisted the best attempts of the machine learning community \cite{ref:deep-learning-article}. 
It also has turned out to be excellent in the discovery of intricate structures in high-dimensional data and therefore a natural candidate for analyzing web data. 
Finally, another advantage of deep learning is that it requires very little feature engineering and can easily take advantage of increases in the amount of available data and computation.

Making predictions based on web data is important for multi-billion dollar online industries, but there is little published research in this area. 
Tech giants such as Google, Facebook and Yahoo published their research some time ago \cite{ref:multimedia-features-yahoo,ref:practical-lessons-facebook,ref:ad-click-prediction-google} but it is now outdated. 
With Zap we had to re-think many aspects of data generation, data collection, feature engineering, applying machine learning models and serving predictions.

% ===============================================================
% ===============================================================

\section{Data Processing}
\label{sec:data-processing}

We start by reviewing the data collection procedure.

Analytics events such as page views, clicks, scrolls, positive outcomes (e.g. purchase), etc. are collected by a JavaScript library. 
Similar events from mobile applications are collected by a Mobile SDK. 
These libraries are comparable to Google Analytics\footnote{\url{https://www.google.com/analytics/}} or Segment\footnote{\url{https://segment.com/}}.

\begin{figure}[h!] % * - image across two columns
\center{\includegraphics[width=0.8\linewidth]{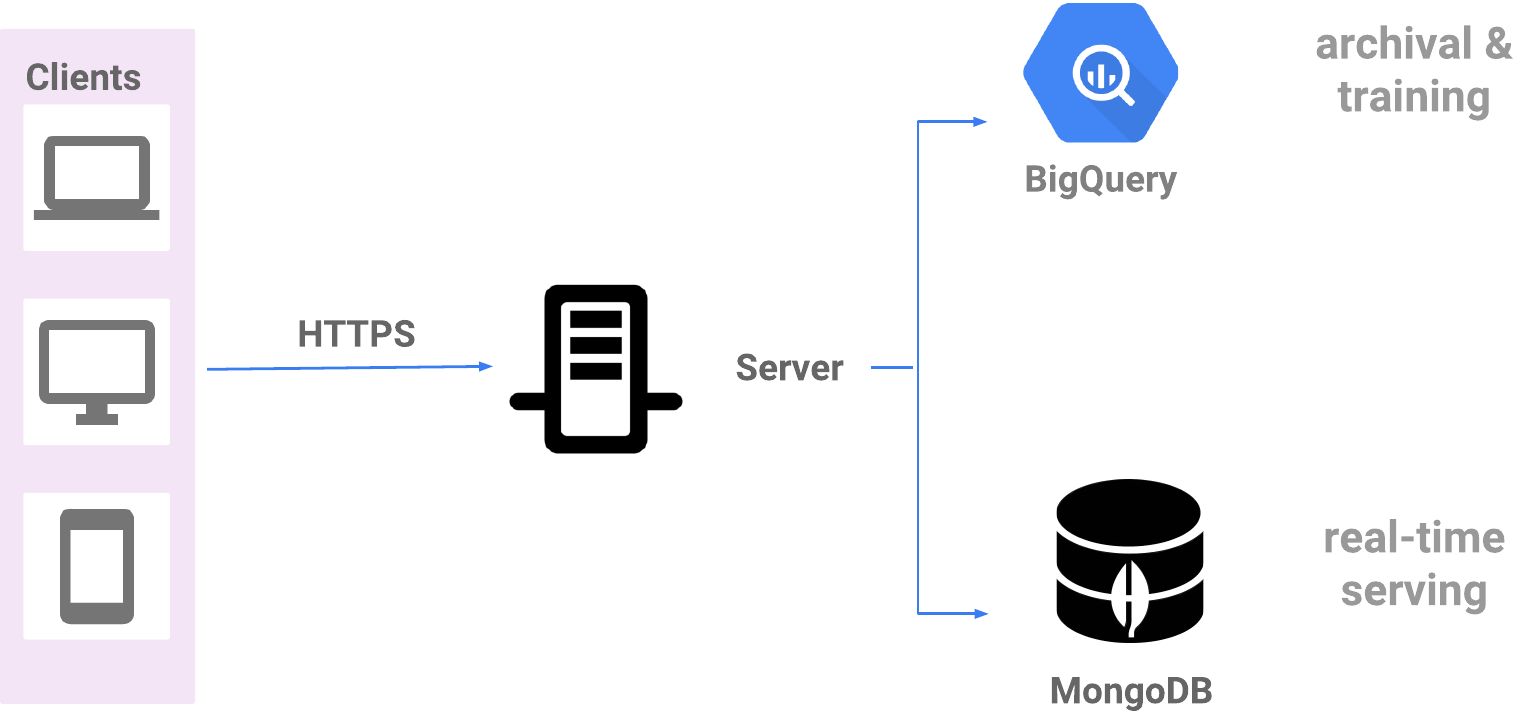}}
\caption{Data collection.}
\label{fig:data-collection}
\end{figure}

Collected data is posted in real-time to our server end-point, either event-by-event or in batches. 
Then the data is cloned into two streams: one that goes to our long-term archival solution (Google BigQuery), and another that goes to a fast database (such as MongoDB or Redis). 
The fast database is used for real-time predictions (see Fig. \ref{fig:data-collection}). 
Only a limited number of the most recent events and a subset of relevant event types are stored in the real-time database, just enough to be able to compute the predictions when the client queries for them. 
Long-term archival databases are used for training, but are not suited to real-time serving. 
Data from the real-time database is typically aged out after a few days, while long-term data is archived in a data warehouse.

Three key concepts used in our approach are \textit{user sessions}, \textit{instances} and \textit{examples}. 
They will be described in details in the next two subsections.

% ===============================================================

\subsection{User Sessions}
\label{subsec:sessions}

A \textit{user session} is a stream of chronological events tracking the actions of a user that occur on a client device such as a web browser or mobile application. 
Each event has a \texttt{timestamp}, an \texttt{anonymousId}, a \texttt{userId} (if the user is logged in), and a payload describing the event. 
Some events are generated in response to a user action (e.g. page visit, click, scroll) while others may be synthetic events (e.g. a prediction point) automatically inserted by the client libraries or by the server.

Data preprocessing performed during serving and training are slightly different (see Fig. \ref{fig:sessions-instances}). 
First, we will describe the process used in training, then discuss how serving is different.

The first stage of our data preprocessing pipeline is transforming events from BigQuery (or another database) into user sessions. 
We run an Apache Beam job that maps rows from BigQuery to \texttt{anonymousId} and then performs a \texttt{GroupByKey} operation. 
Next we filter out events not generated by a real user, this includes events generated by scrapers, bots, our models, testing events, etc., and ignore long sessions\footnote{A large session is an arbitrary and a website specific term, we typically drop user sessions with more than 1000 events.}. 
Finally, forming user sessions is completed by ordering events by timestamps.

Next, we perform what can be thought as preliminary feature engineering: we remove duplicate information from web events, parse URLs and extract user agents. 
The importance of this step will be clarified in Sec. \ref{subsec:feature-extraction}.

% ===============================================================

\subsection{Instances and Examples}
\label{subsec:instances}

\begin{figure*}[h!] % * - image across two columns
\begin{subfigure}{0.45\linewidth}
\center{\includegraphics[width=1\linewidth]{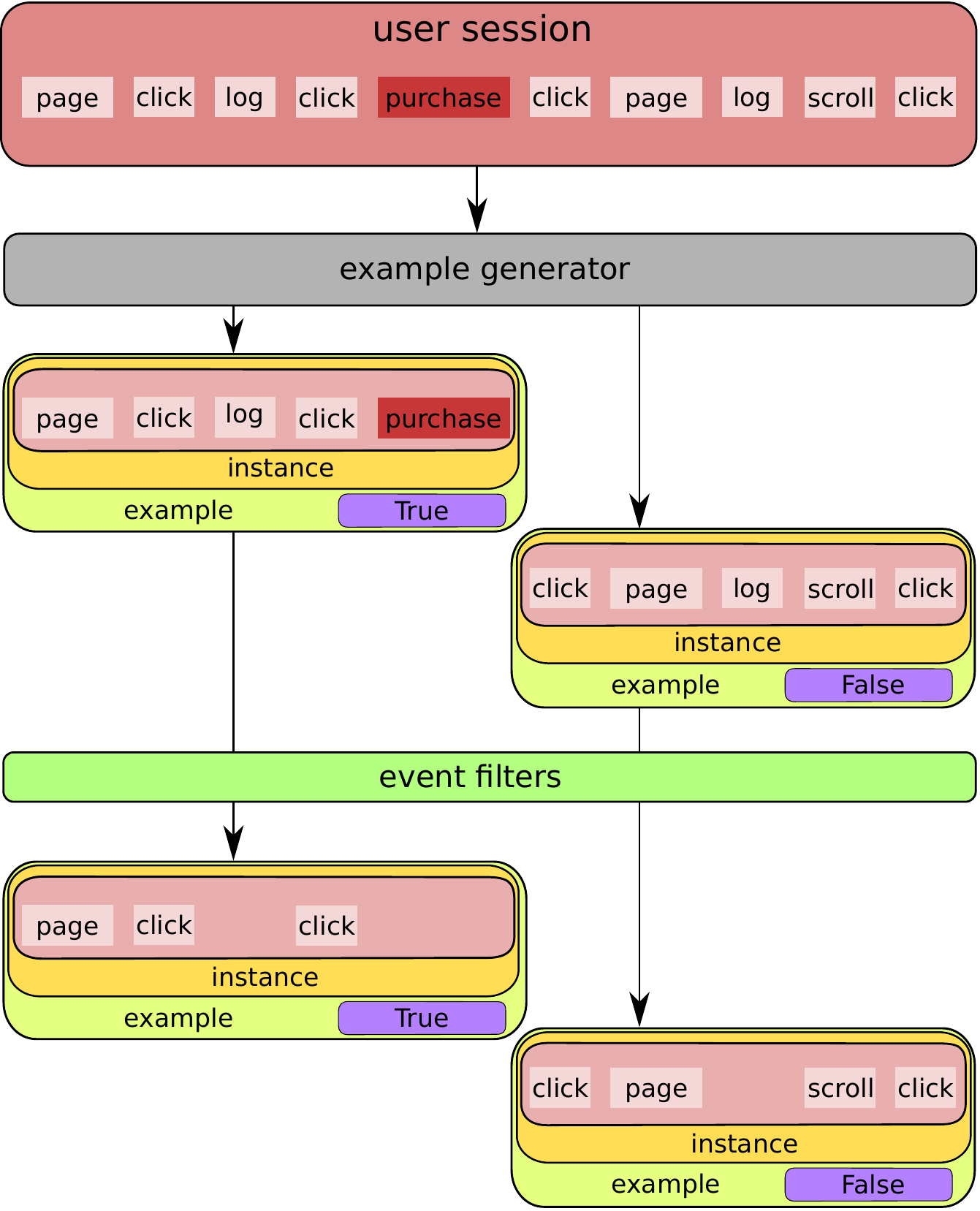}}
\caption{Training}
\label{fig:session-instances-training}
\end{subfigure}
\hfill
\begin{subfigure}{0.45\linewidth}
\center{\includegraphics[width=1\linewidth]{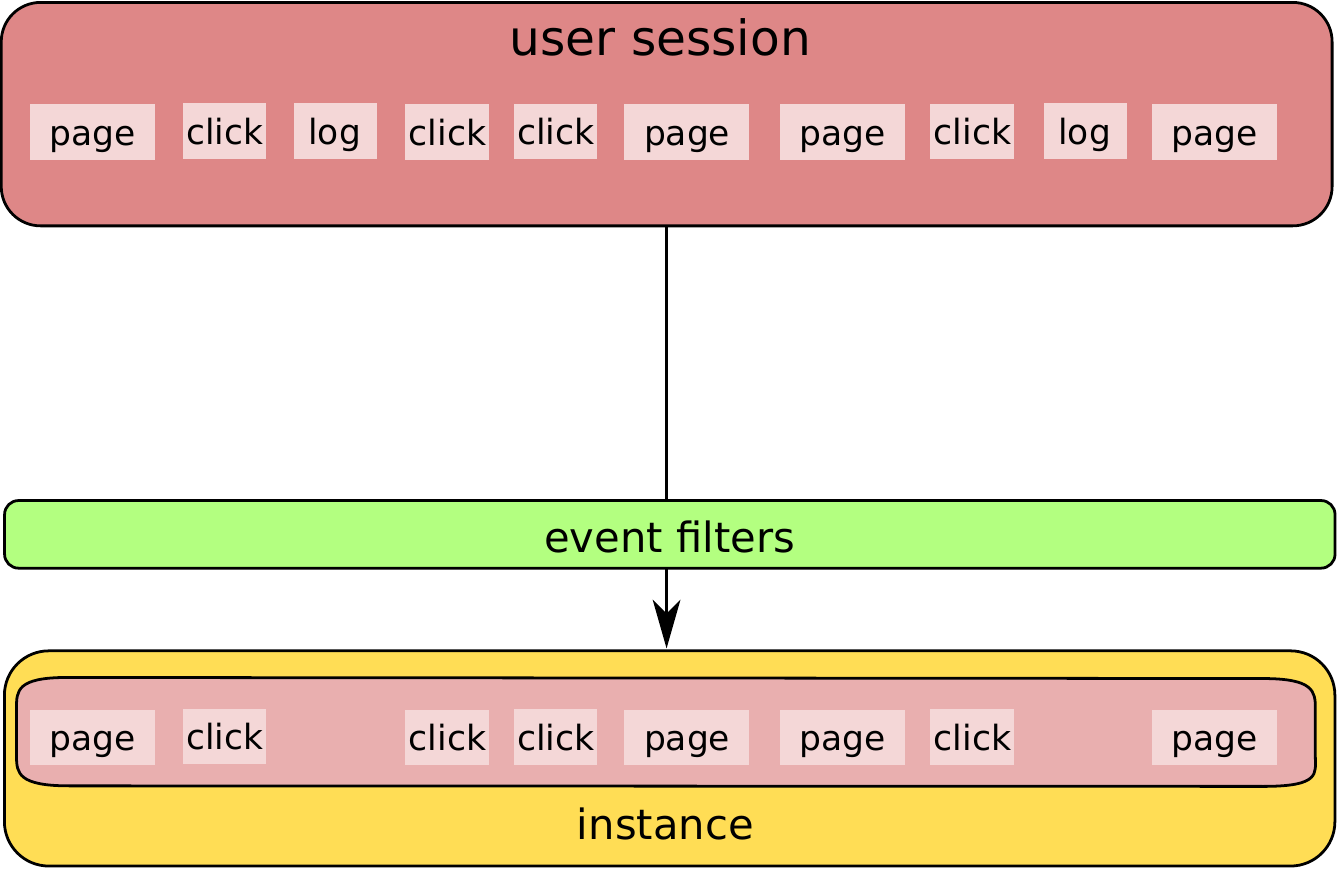}}
\caption{Serving}
\label{fig:session-instance-serving}
\end{subfigure}
\caption{Constructing instances and examples from user sessions in (a) training and (b) serving. 
User session consists of events such as page, click, log, scroll and purchase. 
Example generator picks several events (according to the desired logic), then event filters remove unwanted events such as log events or clicks/scrolls. 
For training, the example generator also generates a label (according to the desired logic) to form an example. 
In serving, the entire session at a given point in time is an instance, so the example generator is not used.}
\label{fig:sessions-instances}
\end{figure*}

After user sessions are constructed, they are sliced into what we call prediction instances (see Fig. \ref{fig:sessions-instances}), which are basically fragments of user sessions generated by an \textit{example generator}. 
An example generator defines points within the user session where predictions would have been made. 
For training, each instance is combined with a label to form an \textit{example}, which is also done by an example generator (see Fig. \ref{fig:session-instances-training}). 
In serving, the example generator is not used (see Fig. \ref{fig:session-instance-serving}). 

A procedure of constructing examples from a session is demonstrated in Fig. \ref{fig:session-instances-training}. 
Depicted user session contains a purchase event, which we call a \textit{positive event}. 
The example generator goes through the session looking for a positive event. 
As soon as it finds one (or the end of the session is reached\footnote{Slicing rules could be different and are specified by a programmer.}), the example generator creates a label. 
Then, \textit{event filters} remove all irrelevant information or information which could lead to bogus predictions to form an instance. 
For example, predictive models could just map a purchase event to the actual purchase, but since the goal is to predict this event based on the data available before it happens, using this event is clearly a bogus behavior.

Finally, by attaching a label to the instance we get an example. 
Examples constructed in this way are then fed into machine learning models\footnote{Note that because we use neural networks as our machine learning models, raw web data has to be transformed into a useful representation before being fed into neural networks. 
This transformation is discussed in the next section Sec. \ref{subsec:feature-extraction}}. 
It should be emphasized that instance generators are the \textit{only} website-specific part of Zap.

Along with instances, some of our machine learning models use \textit{metadata}, which is a characteristic of the user or the session in general. 
This data is not of a sequential nature, so, as will be explained in Sec. \ref{sec:predictive-models}, it is fed into the machine learning models separately (see Fig. \ref{fig:rnn-model}).

% ===============================================================

\subsection{Feature Scaling and Instance Coding}
\label{subsec:feature-extraction}

The last stage of our data preprocessing pipeline is transforming instances and examples into representation suitable for our deep learning models: the input data should be a sequence of real valued feature vectors. 
Moreover, it is a good practice in general and particularly important for our models to perform feature scaling. 
We further restrict feature vectors to have values in $[0, 1]$-range. 
This is done because, as will be explained later in this section, we combine several different types of data such as strings and numbers into one feature vector (see Fig. \ref{fig:web-data}). 
If, however, some elements of this vector are not of the same scale, i.e. features with smaller numerical values, they might be suppressed by other features from a larger scale. 
That could lead to models converging to a worse local minima and, as a consequence, to less accurate model predictions\footnote{Some advanced optimization techniques work better with non-scaled features. 
For example, Adam \cite{ref:adam} maintains different learning rates for each network parameter and separately adapts them as learning unfolds.}.

We hash web data strings, e.g. URLs, user agents, into a real-valued feature vector that we call \textit{hash buckets}\footnote{This approach is inspired by the Bloom filter \cite{ref:bloom}.}. 
Consider a set of strings $S$, which needs to be transformed into a feature vector $v$ (which then will be fed into a neural network). 
We start by initializing $v$ with zeros. 
Next, for $s_i \in S$ we apply a hash function $h$ to $s_i$ as follows $h_i=h(s_i), \bar{h_i}=h(s_i+``some\_fixed\_string")$, finally we set $v[h_i]=1, v[\bar{h}_i]=1$. 
Note that $v$ should have enough capacity to avoid collisions. 
We empirically determined that for a typical website setting $\mathrm{dim}(v)=100$ results in the collision rate of 10\%. 
The sparsity in this case is approximately 80\%\footnote{The detailed analysis of sparsity will be performed in the follow-up paper.}. 
This procedure leads to creation of a bit vector $v$. 
See Algorithm \ref{alg:hash-buckets}.

\begin{algorithm}
\caption{Hash Buckets Generation}
\label{alg:hash-buckets}
\begin{algorithmic}
\REQUIRE A set of strings $S$, a hash function $h$ and an integer $n$. 
\ENSURE Vector $v$, such that $\mathrm{dim}(v)=n$ and $v_i\in[0,1]$.
\STATE Let $v=[0]^n$.
\FORALL{$s_i\in S$}
	\STATE Compute two hash values:\\ $h_i = h(s_i), \bar{h}_i = h(s_i + ``some\_fixed\_string")$.
	\STATE Set bits in $v$ using $h_i$ and $\bar{h}_i$ as indices: \\ $v[h_i] = v[\bar{h}_i] = 1$.
\ENDFOR
\end{algorithmic}
\end{algorithm}

For categorical data we use standard \textit{one hot encoding}: if the number of possible values of a particular variable is limited to a fixed set of length $m$ this variable can be represented as a bit vector, $w$ (of length $m$). 
Categorical values should be mapped to integer values. 
Then each integer value is represented as a bit vector that is all zero values, except the index of the integer which is marked as one.

\begin{figure}[h!] % * - image across two columns
\center{\includegraphics[width=0.8\linewidth]{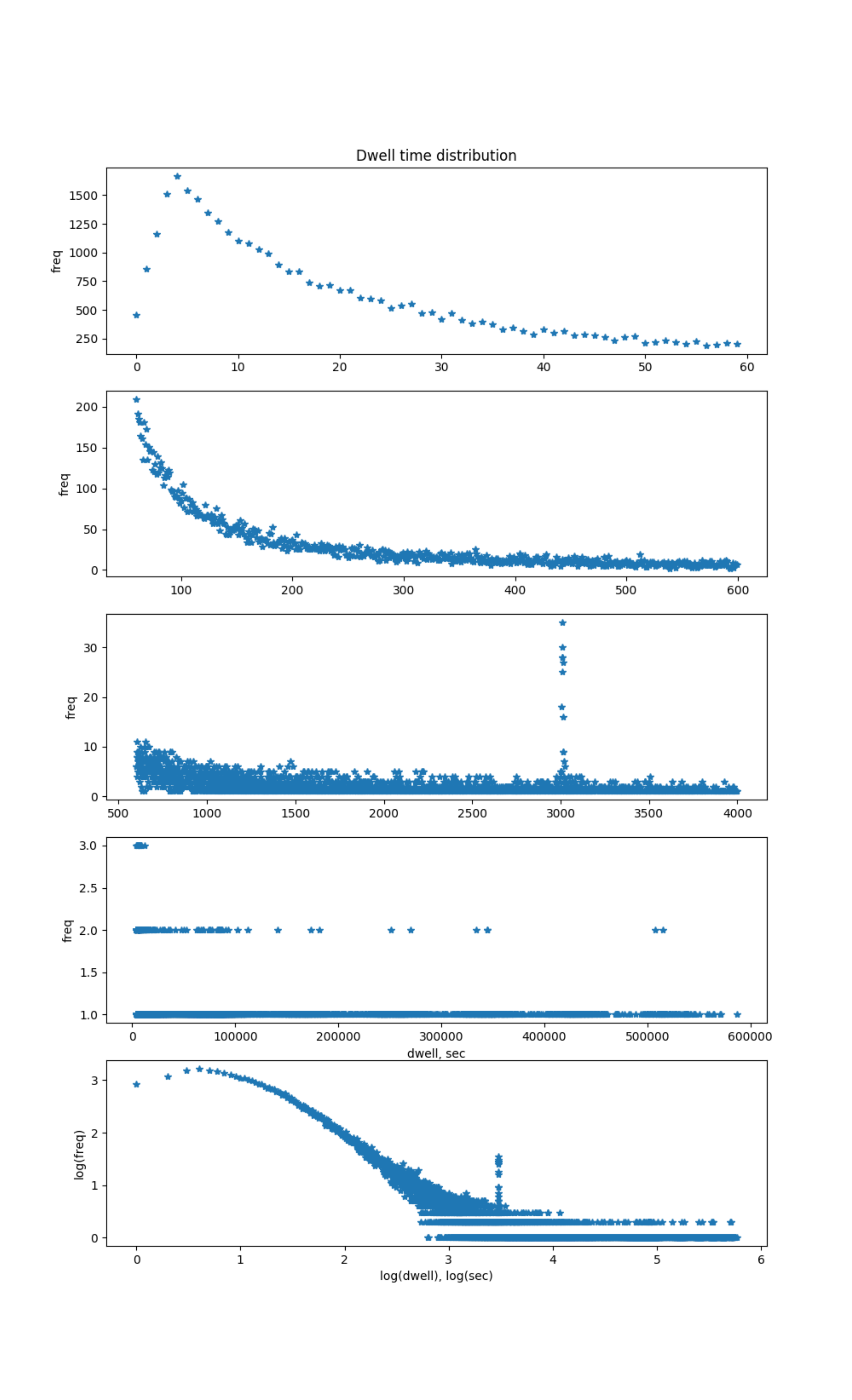}}
\caption{A typical dwell time distribution (distribution of time intervals between page loads measured in seconds). 
The peak at the 1 hour mark is an artifact due to the website refreshing pages after 1 hour of inactivity.}
\label{fig:dwell-time-distr}
\end{figure}

Along with text data we compute numerical features (such as dwell time between events), map them into $[0,1]$-range and then concatenate to other features such as hash buckets, $v$, computed at the previous stage. 
To achieve it we use two different techniques: \textit{normalization} and \textit{data bucketing (binning)}\footnote{Terms ``bucketing" and ``data binning" will be used interchangeably in this article.}. 
If the range of a numerical value, $X$ is known, $X\in[X_{min}, X_{max}]$, one can simply apply normalization as 
\begin{equation}
X' = \frac{X - X_{min}}{X_{max} - X_{min}},
\end{equation}
which ensures that $X'\in[0,1]$. 
If the range is unknown or feature values are distributed non-uniformly, then data binning is preferred. 
A relevant example of a non-uniform distribution is showed in Fig. \ref{fig:dwell-time-distr}, where we plot a typical dwell time distribution (distribution of time intervals between page loads measured in seconds)\footnote{Another example of dwell time distribution can be found in \cite{ref:dwell-dist}.}.

Data binning is a form of quantization, in which original values that fall in a given small interval, a bin, are replaced by a value representative of that interval. 
In our pipeline numerical values are converted into bit vectors. 
For example, rounding a floating point number to an integer and then turning it into a one-hot vector is an example of \textit{linear bucketing}. 

Let's say our feature's current value is 42 and it is distributed in the range $[0, 100]$. 
If we want to bucket it into 11 bins its current value would be transformed into $[0, 0, 0, 1, 0, 0, 0, 0, 0, 0]$, where the first bin corresponds to values in $[0, 10)$-range, the second one in $[10, 20)$ and so on. 
In this example all bins are of the same size (that is why they are called linear bins). 
A simple Python implementation is shown on Listing \ref{code-bucketing}.

\begin{lstlisting}[caption={Linear bucketing with \texttt{numpy}.}, label=code-bucketing, numbers=left, language=Python]
def get_one_hot_bucket_vector(value, start, end, n_buckets):
    step = (end - start)/(n_buckets - 1)
    feature = numpy.zeros(n_buckets)
    buckets = numpy.arange(start, end + 1, step)
    index = numpy.digitize(value, buckets)
    feature[index - 1] = 1
    return feature

n_buckets = 11
start, end = 0, 100
value = 42
bucketed_feature = get_one_hot_bucket_vector(value, start, end, n_buckets)
\end{lstlisting}

\begin{figure}[h!] % * - image across two columns
\begin{subfigure}[h!]{0.35\linewidth}
\center{\includegraphics[width=1\linewidth]{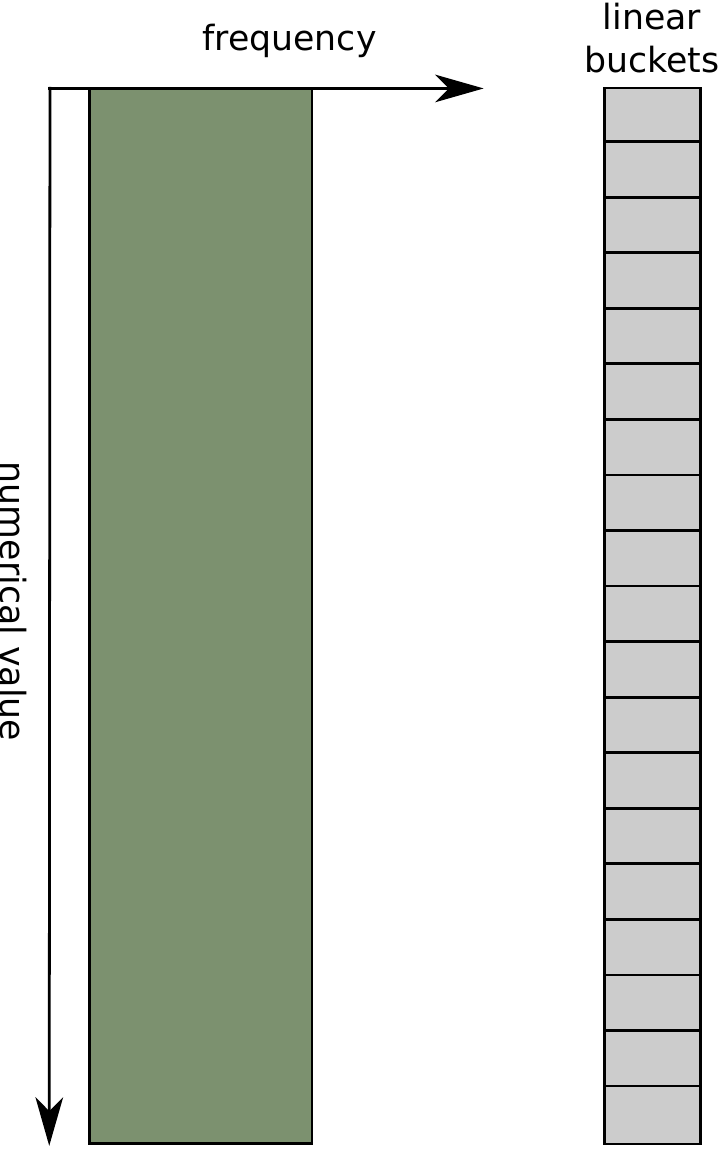}}
\caption{Linear buckets}
\label{fig:linear-buckets}
\end{subfigure}
\hfill
\begin{subfigure}[h!]{0.35\linewidth}
\center{\includegraphics[width=1\linewidth]{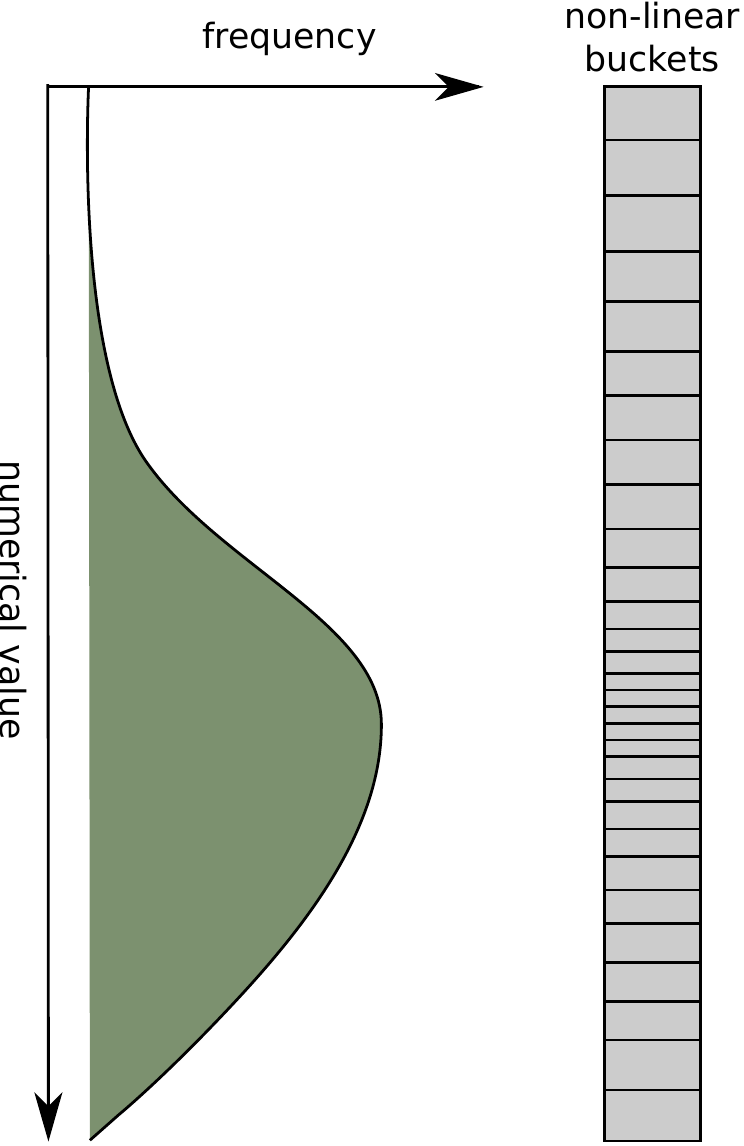}}
\caption{Non-linear buckets}
\label{fig:non-linear-buckets}
\end{subfigure}
\caption{Pictorial representation of linear and non-linear buckets. 
(a) For uniformly distributed data, using buckets of the same size is a good quantization method without much information loss. 
(b) If data distribution is non-linear, linear buckets would result in heavy information loss because most values (which are close to the distribution peak) would be assigned the same value. 
In this case a better approach is to use buckets of different size, larger buckets near the tails and smaller buckets near the peak.}
\label{fig:buckets}
\end{figure}

This technique works well for features with uniformly distributed values as shown in Fig. \ref{fig:linear-buckets}, but would lead to a big information loss when applied to variables with non-uniform distributions. 
A better solution is to have buckets whose sizes depend on the distribution density: it is desirable to have a higher resolution (which corresponds to having more buckets) near the distribution peak, and lower resolution away from the peak (see Fig. \ref{fig:non-linear-buckets}). 

Creating perfect buckets is not a simple task because one needs to know a functional form of the given distribution, which is almost never known a-priori, so for our tasks we assume a simple power law distribution to create more buckets near the peak and less at the tail. 
This approach is a good approximation for data with distributions such as depicted in Fig. \ref{fig:dwell-time-distr}. 
Essentially we combine linear and non-linear buckets. 
To do so we specify three more parameters: linear buckets step, $s_l$, linear buckets cutoff, $c_l$, and non-linear buckets cutoff, $c_n$. 
Then linear, $n_l$, and non-linear, $n_n$ buckets are computed as follows:
\begin{eqnarray}
n_l^i = s_l i, &\quad& 0\le i\le c_l,\\
n_n^i = s_n p^i, &\quad& 1\le i\le N - c_l,
\end{eqnarray}
where $s_n$ and $p$ are determined from the following boundary conditions:
\begin{eqnarray}
n^i_n|_{i = 0} = c_l, \quad n^i_n|_{i = N - c_l} = c_n.
\end{eqnarray}
The first one is a continuity of linear and non-linear buckets, and the second one is the upper bound on non-linear buckets. 
Once buckets are generated, the procedure outlined in Listing \ref{code-bucketing} is applied (except line 3 is replaced with buckets constructed beforehand). 
We also use a more sophisticated technique that allows choosing the desired resolution for non-linear buckets.

% ===============================================================
% ===============================================================

\section{Predictive Models}
\label{sec:predictive-models}

\begin{figure*}[h!] % * - image across two columns
\begin{subfigure}[h!]{0.75\linewidth}
\center{\includegraphics[width=1\linewidth]{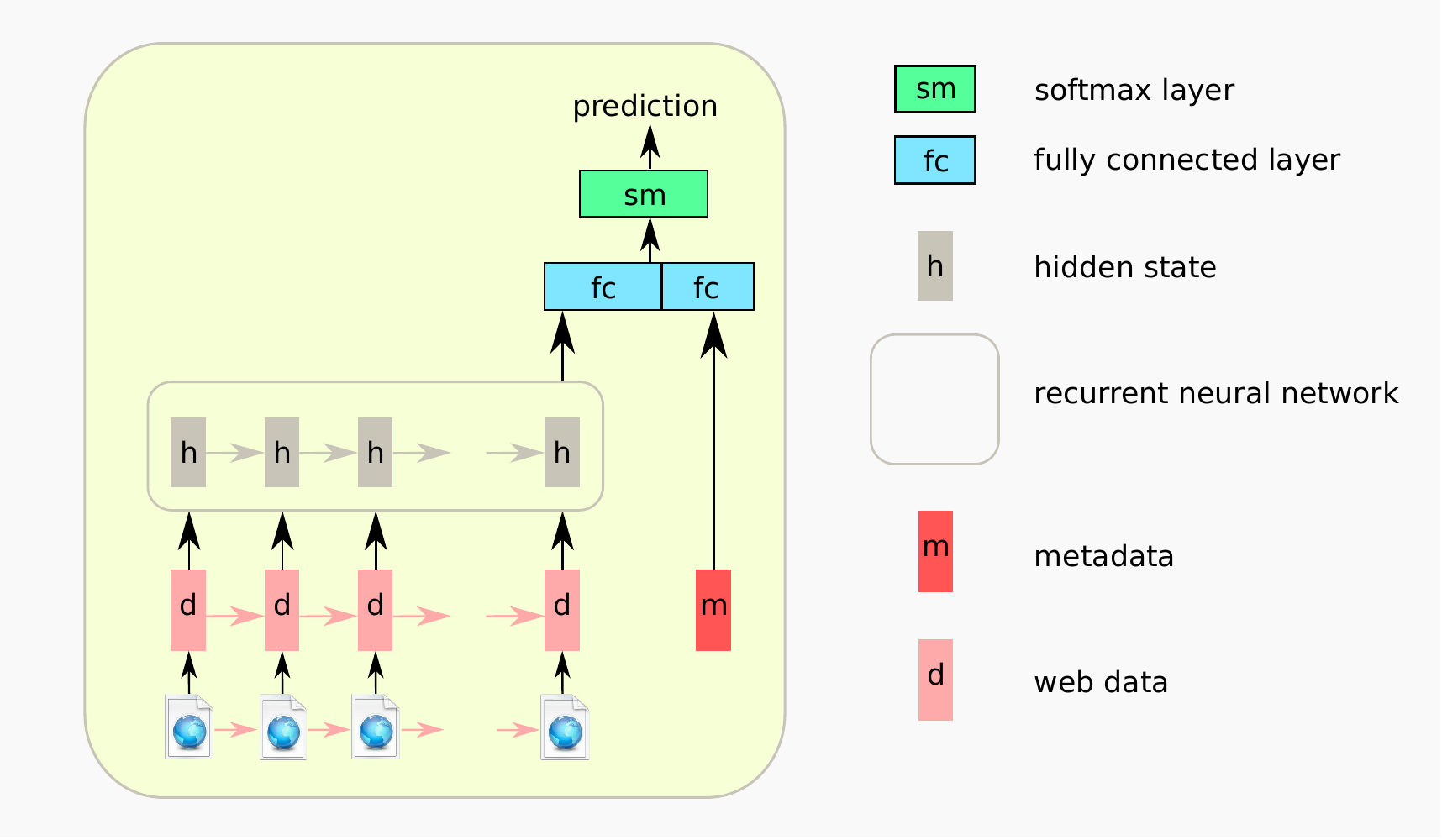}}
\caption{RNN model architecture.}
\label{fig:rnn-model}
\end{subfigure}
\hfill
\begin{subfigure}[h!]{0.15\linewidth}
\center{\includegraphics[width=1\linewidth]{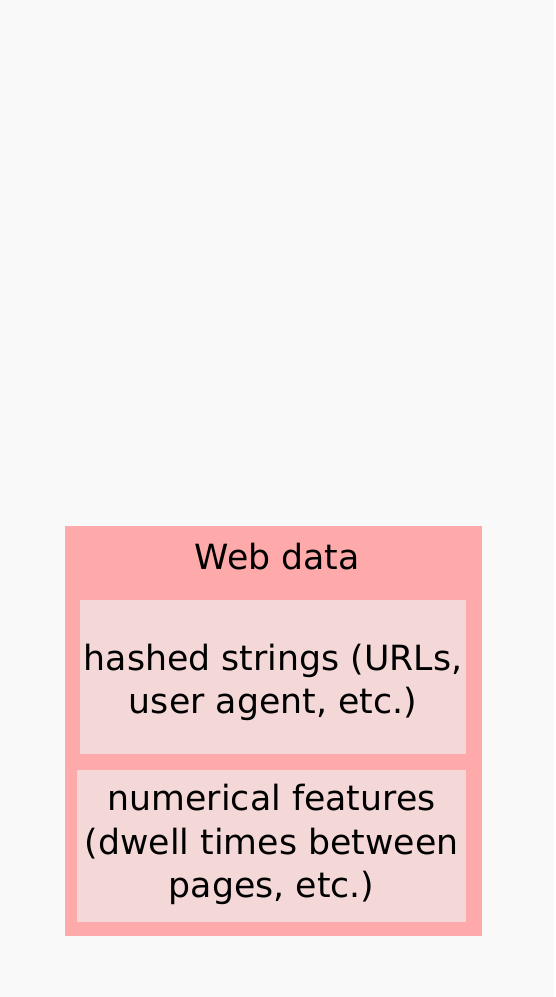}}
\caption{Web data.}
\label{fig:web-data}
\end{subfigure}
\caption{RNN model architecture and data. 
(a) Our model uses an RNN module to process sequential web data (e.g. page visits, clicks, scrolls, etc.) and several fully connected layers to process metadata, which characterizes a user session as a whole. 
(b) Each element of a user session is a combination of hashed strings and numerical features.}
\label{fig:model-data}
\end{figure*}

After preprocessing, the data is fed into predictive models. 
Our models consist of two parts: one processing sequential data and another one processing metadata. 
The sequential data is processed via a recurrent neural network (RNN) or a convolutional neural network (CNN). 
In this paper we focus on the RNN architecture. A CNN approach will be described in a follow-up paper.

The RNN architecture\footnote{As a recurrent unit we use Gated Recurrent Unit (GRU), which showed the best performance in our tests.} is depicted in Fig. \ref{fig:rnn-model}. 
The metadata is processed via several fully-connected layers, which is then merged with the output of RNN and passed to the softmax layer. 
Note that each element of sequential data is obtained by concatenating hash buckets with buckets for numerical features such as dwell time (see Fig. \ref{fig:web-data}), and metadata is fed separately (see Fig. \ref{fig:rnn-model}).

We train our models with the Adam optimizer \cite{ref:adam} and use L2 and dropout \cite{ref:dropout} for regularization. 
Another important component of our training approach is using cross entropy weighting \cite{ref:weighted-cross-entropy}. 
The datasets we work with are highly unbalanced due to the nature of problems we deal with: positive events could be very rare (e.g. $<1\%$), which makes training difficult. 
Weighting cross entropy allows one to trade off recall and precision by up- or down-weighting the cost of a positive error relative to a negative error. 
Another way to deal with highly unbalanced datasets is to use negative down-sampling. 
However since we also want our predictions to reflect the actual probabilities (as will be explained in Sec. \ref{subsec:recalibration}) we do not use negative down-sampling.

Finally, to find the best model we do a hyperparameter search over L2, dropout and the number of RNN units.

% ===============================================================
% ===============================================================

\section{Real-time Serving of Trained Models}
\label{sec:serving}

As mentioned in Section \ref{sec:data-processing}, incoming client data is cloned into two streams: one for long term archival and training, another one for real-time serving, stored in MongoDB (see Fig. \ref{fig:data-collection}).

\begin{figure}[h!] % * - image across two columns
\center{\includegraphics[width=0.6\linewidth]{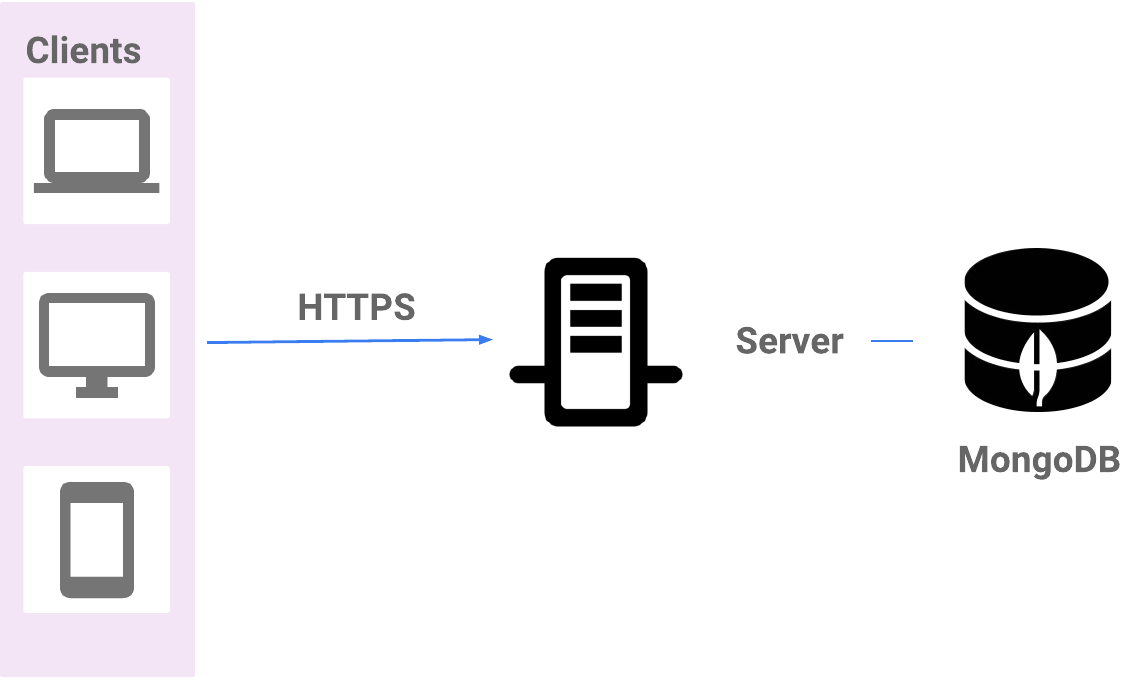}}
\caption{Client API call to retrieve a prediction. The server queries MongoDB for events and inferences using frozen models loaded in memory.}
\label{fig:api-prediction}
\end{figure}

Model predictions are served via a REST API. 
The client making the API call can be a web application (JavaScript) or a mobile application. 
When a request comes in, the list of most recent events for the given user ID is pulled from MongoDB, ordered by event timestamp. 
The size of the list depends on the model fixed input size, e.g. 40 most recent events (if not enough events are available, the list is padded with blank events up to the fixed size). 
The events are turned into an instance and run through the currently active model version to generate a prediction. 
The prediction value is returned to the client application which can then take an action based on the value. 
This procedure is demonstrated in Fig. \ref{fig:api-prediction}.

For each application, there is a current version of the model loaded in the server memory. 
The models are stored and loaded in memory in a TensorFlow frozen form which speeds up loading and reduces memory footprint.

% ===============================================================
% ===============================================================

\section{Re-training Models}
\label{sec:retraining}

Since our models make predictions for a particular website or app, it is important to keep the models up-to-date with the current version of that website or app.

New model versions are re-trained automatically on a periodic basis, e.g. weekly. 
This ensures that the latest collected data is used and models take into account the most recent changes to application structure and user behavior. 
There is an automated process that goes through the following steps every week:
\begin{itemize}
\item Re-trains a new model for each application as described in the previous sections. A rolling window of most recent data is used, e.g. 7-60 days, depending on how much data each application produces.
\item Validates that the new model is acceptable by checking certain performance metrics such as AUC (see more about metrics used in Section \ref{sec:evaluation-of-model-performance}) vs. previous model versions in the same family.
\item Archives the new model version together with stats about it and the examples generated during training (see Section \ref{subsec:instances}). This allows reverting back to an older version of a model if issues are discovered in production, provides a detailed history of how the model evolved and allows comparing different versions and their stats.
\item Verifies that the new model will be served correctly. During training, a random subset of examples (e.g. 1000) is saved together with the computed prediction value for each example. The verification consists of running these examples through the exact code path used for real-time serving with the expectation that the served prediction value exactly matches the value computed during training. This ensures that the same common code is used both in training and serving (see Fig. \ref{fig:sessions-instances}).
\item If validation and verification pass, the new model version is auto-deployed to production and new predictions are served using the new version.
\end{itemize}

% ===============================================================
% ===============================================================

\section{Post Serving Analysis}
\label{sec:post-serving-analysis}

During serving, each prediction made by a model is logged, together with sufficient metadata that can identify the model version and the events used to construct the instance. 
This allows for later analysis of actual model performance by looking at what the model predicted for a user over time vs the actual action the user performed.

% ===============================================================
% ===============================================================

\section{Confidence and Calibrating Predictions}
\label{sec:calibration}

In our applications, it is important to classify user behavior, but it is also important to quantify the expected accuracy of the prediction. 
This problem of predicting probability as a true correctness likelihood is known as \textit{confidence calibration} (for recent developments see \cite{ref:calibration}). 
The importance of calibration is also discussed in \cite{ref:ad-click-prediction-google,ref:practical-lessons-facebook}. 
A rough estimate of how well a model is calibrated is given by \textit{expected calibration error} (ECE). 
In order to compute ECE\footnote{Here we follow \cite{ref:calibration}.}, we group predictions into $M$ interval bins (each of size $1/M$) and calculate accuracy/confidence in each bin. 
Let $B_m$ be the set of indices of samples whose prediction falls into the interval $I_m = (\frac{m-1}{M}, \frac{m}{M}]$, the accuracy is then
\begin{equation}\label{eq:accuracy}
\mathrm{acc}(B_m) = \frac{1}{|B_m|}\sum_{i\in B_m} \mathbf{1}(\hat{y}_i = y_i),
\end{equation} 
where $|B_m|$ is the number of elements in $B_m$, and $\hat{y}_i$ and $y_i$ are the predicted and true class labels for sample $i$. 
The confidence is defined as:
\begin{equation}\label{eq:confidence}
\mathrm{conf}(B_m) = \frac{1}{|B_m|}\sum_{i\in B_m}\hat{p_i},
\end{equation}
where $\hat{p}_i$ is the predicted probability. 
Finally, the expected calibration error is
\begin{equation}
ECE = \sum_{m=1}^M \frac{|B_m|}{n}\left| \mathrm{acc}(B_m) - \mathrm{conf}(B_m) \right|,
\end{equation}
where $n$ is the number of samples. A perfectly calibrated model will have $\mathrm{acc}(B_m) = \mathrm{conf}(B_m)$ for all $m\in M$, so $ECE_{perfect}=0$.

\begin{figure}[h!] % * - image across two columns
\begin{subfigure}[h!]{0.45\linewidth}
\center{\includegraphics[width=1\linewidth]{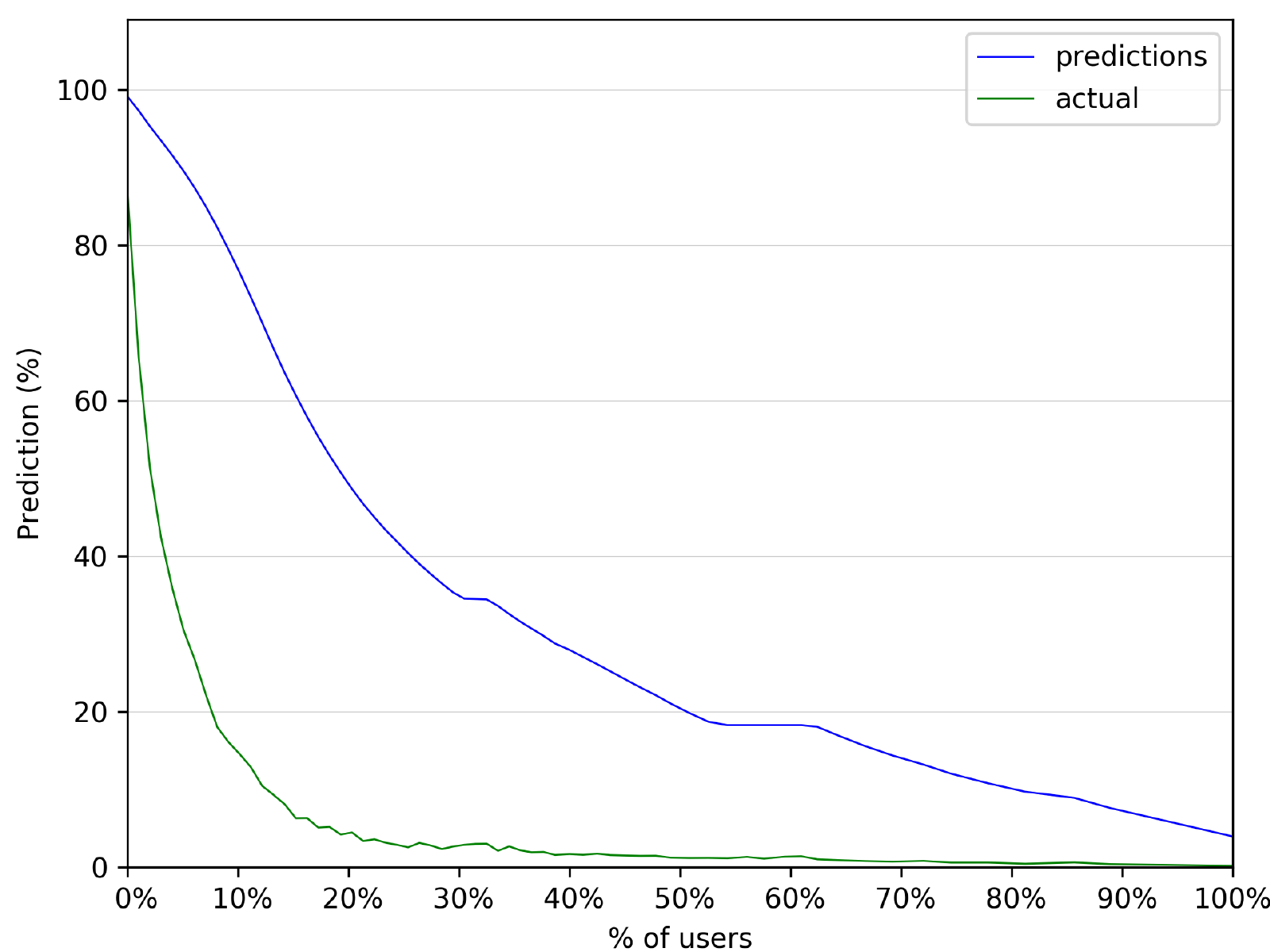}}
\caption{Uncalibrated model.}
\label{fig:uncalibrated-model}
\end{subfigure}
\hfill
\begin{subfigure}[h!]{0.45\linewidth}
\center{\includegraphics[width=1\linewidth]{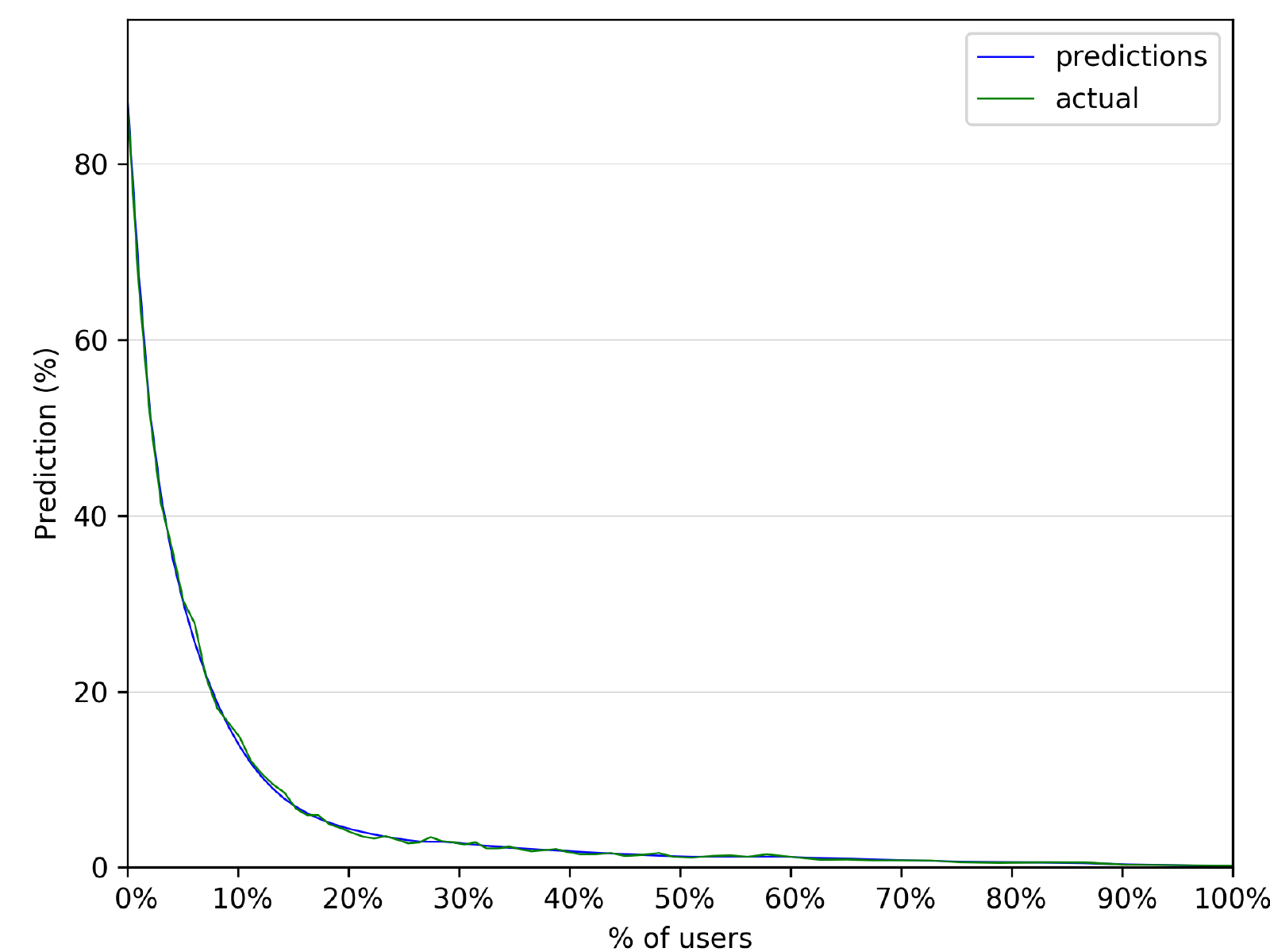}}
\caption{Calibrated model.}
\label{fig:calibrated-model}
\end{subfigure}
\caption{Examples of calibration curves.}
\label{fig:calibration-curves}
\end{figure}

ECE gives only a rough estimate of how well a model is calibrated. 
To get a better understanding, we visualize predictions and actual outcomes by plotting a \textit{calibration curve}. 
We start by sorting training examples by probability from high to low, then we put them into buckets of different size such that each bucket has at least 100 positive outcomes (to minimize noise and maximize resolution). 
Next, we compute confidence \eqref{eq:confidence} and the ratio of actual positive outcomes to the total number of examples in each bucket. Confidence forms a ``predictions" curve, and the ratio of positive outcomes forms an ``actual" curve. 
An example of a calibration curve is depicted in Fig. \ref{fig:uncalibrated-model}. 
We use this graph as a performance metric as will be explained in Sec. \ref{sec:evaluation-of-model-performance}. 
Note that for a perfectly calibrated model the two curves should coincide (see Fig. \ref{fig:calibrated-model}).

We observe that our models turn out to be miscalibrated, a typical calibration curve looks like in Fig. \ref{fig:uncalibrated-model}. 
Possible causes of miscalibration include big model capacity and lack of regularization\cite{ref:calibration}. 
An investigation into our applications shows that cross entropy weighting is the main source of miscalibration. 
This is because we usually deal with highly unbalanced datasets.

% ===============================================================
% ===============================================================

\subsection{Recalibration}
\label{subsec:recalibration}

Since our models turn out to be miscalibrated we perform a \textit{post-processing calibration}. 
There are several methods of doing so (as reviewed in \cite{ref:calibration}), but we found that the best results were obtained by using the \textit{matrix scaling} method. 
The idea here is to apply a linear transformation to the logit vector, $z_i$, produced before the softmax layer for input $x_i$. 
Then the output of a new softmax layer, $\hat{q}_i$, is:
\begin{equation}
\hat{q}_i = \underset{k}{\mathrm{max}}~\sigma_{SM} (\mathbf{Wz_i} + \mathbf{b})^{(k)}.
\end{equation}
The parameters $\mathbf{W}$ and $\mathbf{b}$ are optimized with respect to the error function on a half of the validation set used for training, another half is used for the final validation. 
We use standard gradient descent with a learning rate decay. Fig. \ref{fig:uncalibrated-model} shows a calibration curve for an uncalibrated model, the result after applying the calibration procedure it depicted in Fig. \ref{fig:calibrated-model}.

% ===============================================================
% ===============================================================

\section{Evaluation of Model Performance}
\label{sec:evaluation-of-model-performance}

Most of our models are binary classifiers, so we use AUC (area under the ROC curve) as our main performance metric. 
We also found useful two more metrics, distribution of predictions and calibration curve.

\begin{figure}[h!] % * - image across two columns
\center{\includegraphics[width=0.8\linewidth]{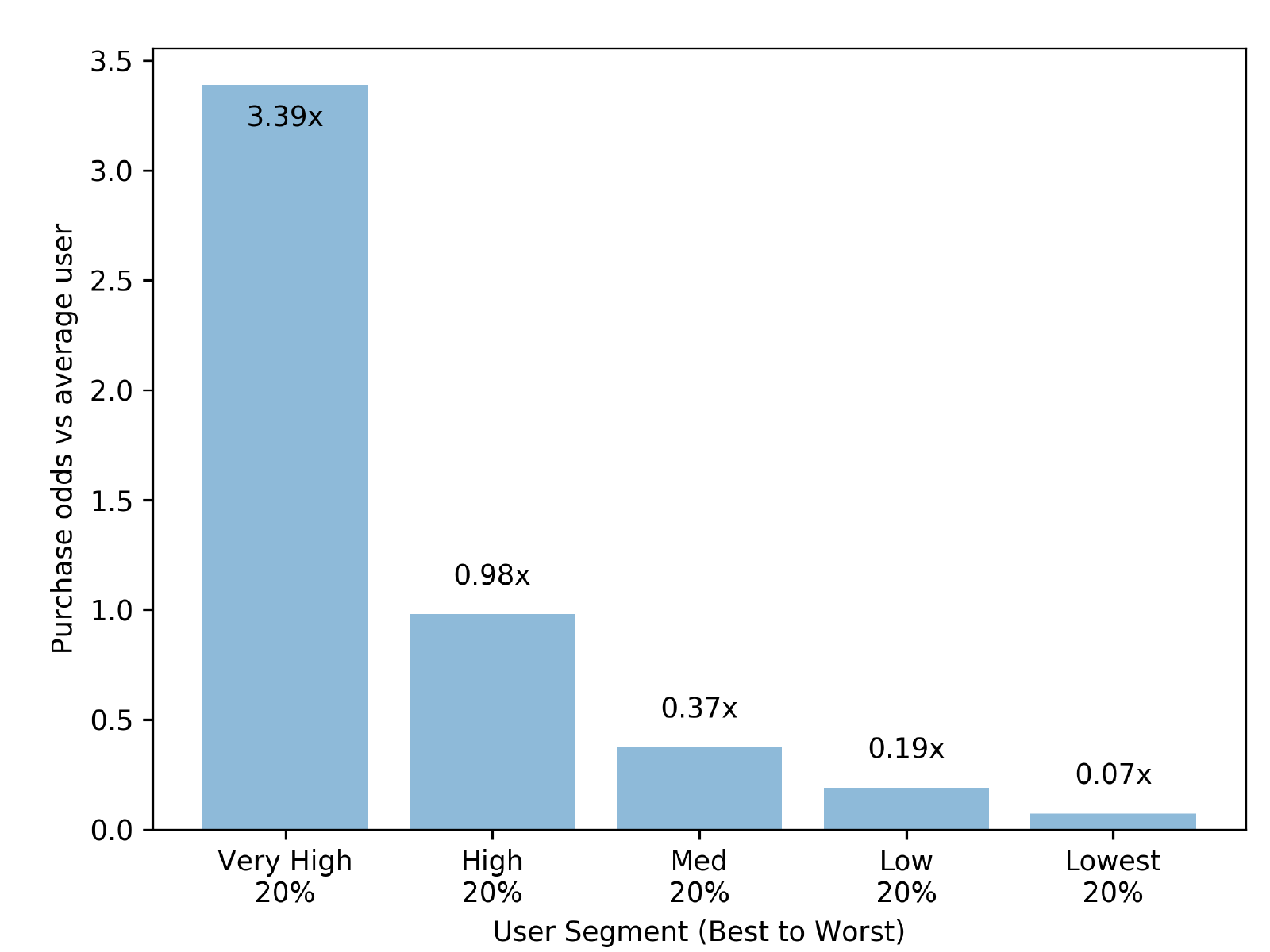}}
\caption{Our predictive models split users into segments with respect to how likely they perform a ``positive" action such a purchase. 
This chart shows that among all users who performed a ``positive" action our model was able to put users who are 3.39 times are more likely to purchase than an average user into the first bucket.}
\label{fig:user-segments}
\end{figure}

Since our models are used to categorize users into several segments for a subsequent action (such as showing users different context, ad retargeting, etc.), we can use the relative distribution of users between different segments as another performance metric. 
To construct segments we first compute the average positive action rate, then we sort predictions from high to low, split them evenly into several\footnote{For our purposes, we split users into 5 segments.} buckets and compute a positive action rate in each bucket, finally we plot buckets and print relative odds of a positive action in each bucket. 
The odds ratio plot reflects how likely the average user in each bucket is to perform a positive action with respect to the base rate.
A typical odds segments plot is depicted in Fig. \ref{fig:user-segments}.
It shows that this model puts users who are 3.39 times more likely to perform a positive event than an average user into a ``very high" bucket, the next bucket has users who are 0.98 times more likely to perform a positive events, and so on.
The ideal model would put all users who generated a positive event into the first bucket\footnote{Unless more than 20\% of users generated positive outcomes, in which case the second bucket will be also used.} and the worst model would distribute users into the buckets evenly.

%One can see that 66.6\% of all users who generated a positive event were identified by a model as ``very high" users (and put into the first bucket), 17\% of users who were identified as ``high" (and put into the second bucket) also generated a positive event, and so on. 
%The ideal model would put all users who generated a positive event into the first bucket\footnote{Unless more than 20\% of users generated positive outcomes, in which case the second bucket will be also used.} and the worst model would distribute users into the buckets evenly.

%\begin{figure}[h!] % * - image across two columns
%\center{\includegraphics[width=0.8\linewidth]{figures/user-segments.png}}
%\caption{Our predictive models split users into segments with respect to how likely they perform a ``positive" action such a purchase. 
%This chart shows that among all users who performed a ``positive" action our model was able to put 66.6\% of those users into the first bucket, 17\% in the second one and so on.}
%\label{fig:user-segments}
%\end{figure}

Finally, as another performance metric we use calibration curve described in Sec. \ref{sec:calibration}. 
Examples of an uncalibrated and calibrated models are given in Fig. \ref{fig:uncalibrated-model} and Fig. \ref{fig:calibrated-model} respectively.

% ===============================================================
% ===============================================================

\section{Interpretability of Predictions}
\label{sec:prediction-texture}

Interpretability of machine learning models is critical for many practical applications. 
First of all, it ensures that the model is aligned with the problem one wants to solve. 
Second, it helps indicate patterns the model actually found in the data so one can use the model's predictions more effectively.

One important example comes from using machine learning in health care: if a model makes a prediction that a patient is at a high risk of an illness, it is very important to help the doctor understand why before ordering expensive or risky treatment. 
Consider another example, if a model identified that a user is going to end her subscription for an online service, it would be very helpful to know why and be able to improve the service quality without having to explicitly ask for feedback. 

It is well-known that deep learning models are hard to interpret due to the large number of parameters and the complex approach to extracting and combining features. 
As this class of models is able to obtain state-of-the-art performance on a wide variety of tasks, more research is focused on linking model predictions to the inputs. 
Some progress has been done in computer vision where the gradients of the target concept calculated in a backward pass are used to produce a map that highlights the important regions in the input for predicting the target concept (e.g. see \cite{ref:pred-vis}). 
This way, one can actually see what a model is focused on when making predictions. 
In our case, it is not applicable, so we follow another path.

Our method is more manual and requires some intuition about the problem at hand and the data in general. 
The idea is the following: while generating examples, we compute statistics that we think are important and store them in each example. 
Then we make hypotheses and use our performance metrics to check them. 
This method is inspired by feature importance analysis, where one only uses a subset of features to see which one is the most important. 
In the case of deep learning, it is not practical since it would require expensive re-training after each choice of features.

\begin{figure*}[h!] % * - image across two columns
\begin{subfigure}[h!]{0.45\linewidth}
\center{\includegraphics[width=1\linewidth]{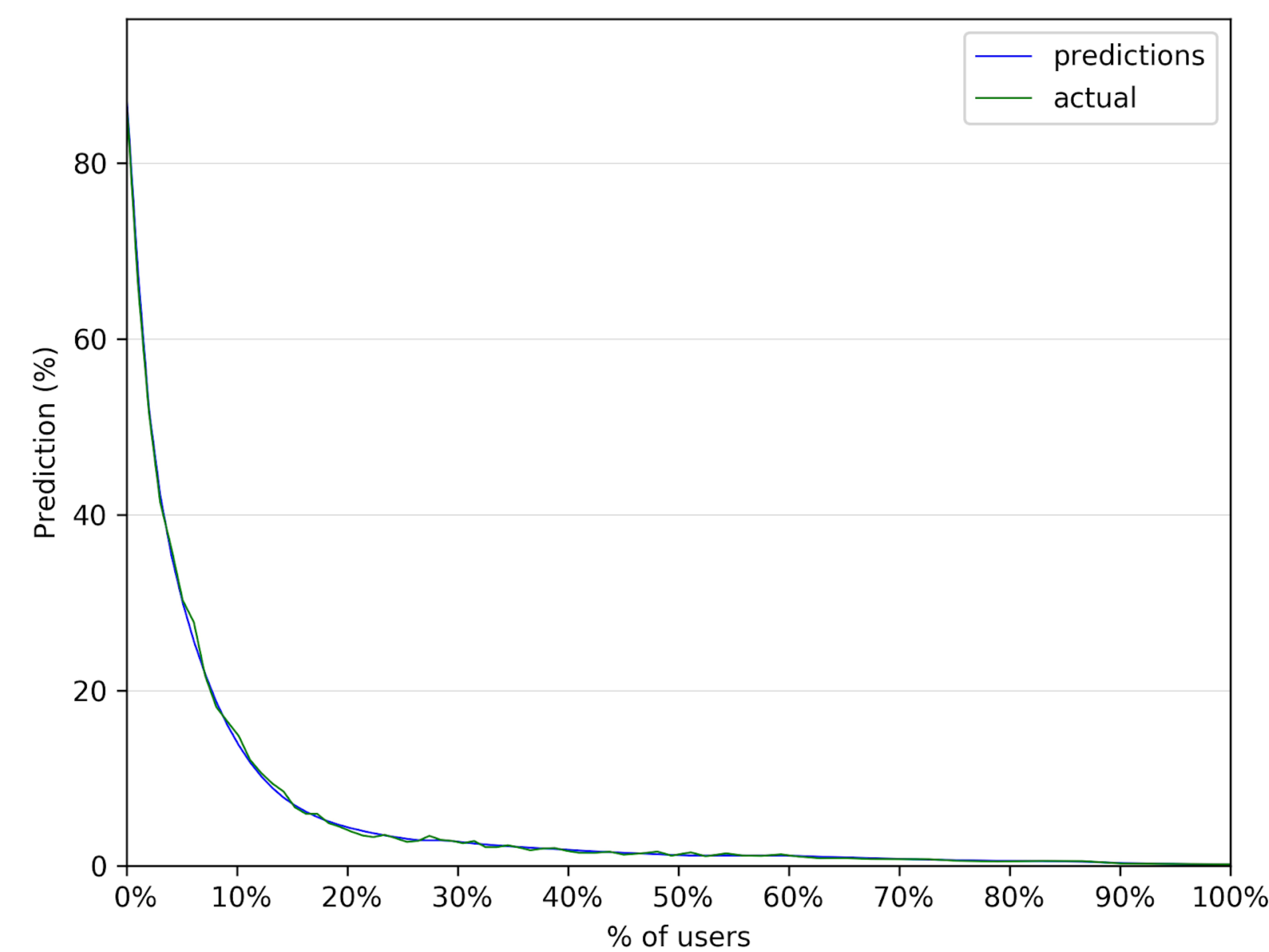}}
\caption{All users.}
\label{fig:prediction-texture-all-users}
\end{subfigure}
\hfill
\begin{subfigure}[h!]{0.45\linewidth}
\center{\includegraphics[width=1\linewidth]{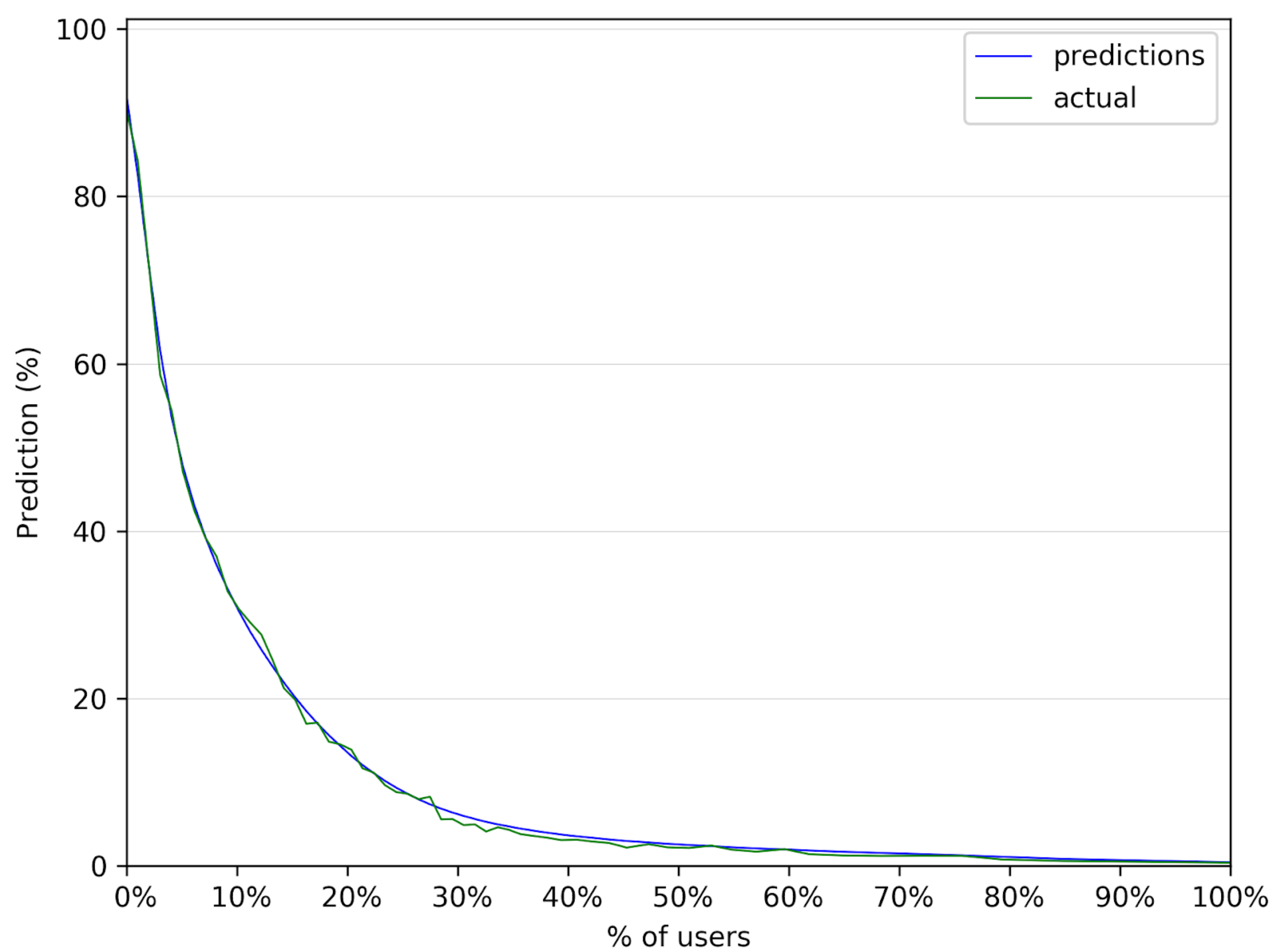}}
\caption{Users with more than 10 clicks who spent more than 1 minute on the website.}
\label{fig:prediction-texture-subset-users}
\end{subfigure}
\caption{The meaning of predictions is extracted from calibration curves. 
Calibration curves for all users (a) and for users with more than 10 clicks who spent more than 1 minute on the website (b) demonstrate that the model is more confident about the latter segment of users, which means that the number of clicks and the time spent on the website are important features as expected.}
\label{fig:prediction-texture}
\end{figure*}

Let's demonstrate the concept via a simple example. 
We have a website with some defined positive event, such as a purchase, and we want to confirm that our model makes reasonable predictions. 
Among all the features fed into our model, let's say we assume that the number of clicks and the time spent on the website are most important. 
To explore this conjecture, we plot calibration curves for users with different number of clicks and different time spent on the website, and compute the positive rate for top 10\% predictions. 
It turns out that the positive rate in top 10\% predictions for \textit{all} users is 0.4, while for users with more than 10 clicks who spent more than 1 minute of the website the positive rate turns out to be 0.6. 
We conclude that the model works according to our intuition. 
The calibration curves are depicted in Fig. \ref{fig:prediction-texture}. 
Of course, the model's logic is more complicated than that, but its interpretability is still an active area of research.

% ===============================================================
% ===============================================================

\section{Use Cases}
\label{sec:use-cases}

Zap is a general pipeline for predicting online user behavior. 
As a predictive analytics tool, we believe it has applicability to a wide range of problems. 
Zap can answer questions about customer intent and audience segmentation. 
How likely is this user to purchase? 
How likely is this user to return to the site? 
Which users are the most valuable long-term user?

Here are a few examples of Zap running in production to solve real-world problems:
\begin{enumerate}
\item Improve ads retargeting by segmenting customers based on likelihood to purchase.
\item Identify customers most likely to join an email list. 
A model is being used to target only high propensity user to avoid spamming all website visitors. 
This model also increases sign-up rates by identifying the best time to ask the user.
\item Identify the most ``in-market" segment of customers. 
This is used to target sales effort on the most promising leads.
\end{enumerate}

We have found that these production models often exhibit a 20/80 type of performance - identifying the 20\% of users who account for 80\% of a desired behavior. 
This performance drives big improvements in the task, whether it is ads ROI or email sign up rates.

% ===============================================================
% ===============================================================

\section{Conclusions}
\label{sec:conclusions}

In this paper we described Zap, a machine learning pipeline that makes predictions based on online user data. 
We designed Zap to process enterprise-level data in order to produce individual predictions that are 1) consumable through an API and 2) understandable to humans. 
The Zap platform goes beyond well-defined academic examples such as image recognition: it turns a marketing or sales question into an actionable decision. 
Furthermore, it provides an automation tool that collects the data and creates the decision algorithm based on that data. 
This is just the first step in building commercial-grade machine learning applications.

Zap is heavily based on Google Cloud Platform technologies such as BigQuery, Cloud Storage, Cloud Dataflow, Cloud Machine Learning, as well as other tools such as MongoDB. 
Big data is processed via Apache Beam and the machine learning component is written in TensorFlow. 
The key feature of Zap is that one can create and serve machine learning models for different websites by specifying example generators which typically consist of a few lines of code. 
The data is generated by our JavaScript library and then cloned into two streams: BigQuery for a long-term archival and MongoDB, which is used to serve real-time predictions. 
To train a new model one uses data from BigQuery, forms user sessions and then examples (which are generated by the only website-specific code in our pipeline called example generators). 
The examples are then fed into a deep learning model, which is trained on Cloud Machine Learning Engine. 
Once a model is trained it can be used to serve real-time prediction using the data coming from MongoDB.

% ===============================================================
% ===============================================================

\section*{Acknowledgments}

The original Zap pipeline was implemented by Josh Sacks and Chris Atenasio. We also thank Mark Vandevoorde and Mike Chu for valuable discussions and feedback on this work.

{\small
\bibliography{references}{}
\bibliographystyle{plain}
}

\end{document}